\documentclass[unnumsec,webpdf,contemporary,large]{oup-authoring-template}%

\graphicspath{{Fig/}}

\usepackage{amsmath,amssymb}
\usepackage{booktabs}
\usepackage{tabularx}
\usepackage{multirow}
\usepackage{makecell}
\usepackage{array}
\usepackage[notables,nolists,nomarkers]{endfloat}

\newcolumntype{Y}{>{\raggedright\arraybackslash}X}

\newcommand{\method}{MOLAR}
\newcommand{\datasets}{MF-PCBA-Noisy7}

\theoremstyle{thmstyleone}%
\theoremstyle{thmstyletwo}%
\theoremstyle{thmstylethree}%

\begin{document}

\journaltitle{xxx}
\DOI{xxx}
\copyrightyear{xxx}
\pubyear{xxx}
\vol{}
\issue{}
\access{xxx}
\appnotes{Paper}

\firstpage{1}

\subtitle{}

\title[Learning Multimodal Molecular Representations from Noisy Labels]{\method{}: Learning Multimodal Molecular Representations from Noisy Labels}

\author[1,4,$\ast$]{Yingxu Wang}
\author[2,$\ast$]{Kunyu Zhang}
\author[3]{Nan Yin}
\author[1,4]{Yu Li}
\author[1,5,$\dagger$]{Eran Segal}

\address[1]{\orgdiv{Department of Machine Learning}, \orgname{Mohamed bin Zayed University of Artificial Intelligence}, \orgaddress{\street{AI Diyafah St}, \postcode{7909}, \state{Abu Dhabi}, \country{United Arab Emirates}}}

\address[2]{\orgdiv{International College}, \orgname{Zhengzhou University}, \orgaddress{\street{Daxue North Road}, \postcode{450000}, \state{Henan}, \country{China}}}

\address[3]{\orgname{The Education University of Hong Kong}, \orgaddress{\state{Hong Kong}, \country{China}}}

\address[4]{\orgdiv{Department of Computer Science and Engineering}, \orgname{The Chinese University of Hong Kong}, \orgaddress{\state{Hong Kong}, \country{China}}}

\address[5]{\orgdiv{Department of Molecular Cell Biology}, \orgname{Weizmann Institute of Science}, \orgaddress{\state{Rehovot}, \country{Israel}}}

\corresp[$\ast$]{Equal Contributions.}

\corresp[$\dagger$]{Corresponding author: \href{mailto:Eran.Segal@weizmann.ac.il}{Eran.Segal@weizmann.ac.il}}

\abstract{
\textbf{Motivation:}
Noisy labels are a common challenge in molecular property prediction because molecular annotations are often obtained from assays, curated databases, or weak annotation pipelines rather than directly observed clean biological states. Treating recorded labels as reliable supervision can cause models to memorize corrupted observations and learn misleading molecular evidence. In multimodal molecular representation learning, this issue can be amplified by graph--text fusion or alignment, which may propagate label-induced errors across modalities. \\
\textbf{Results:}
We propose \method{}, a noise-aware framework for learning multimodal molecular representations from noisy labels. \method{} separates latent clean-property inference from recorded-label observation: graph and text views contribute residual evidence to a clean-property distribution, and a categorical label-observation channel maps this distribution to recorded labels for training. This formulation derives posterior label reliability and modality-specific molecular evidence from the model. Experiments on naturally noisy molecular benchmarks and controlled label-flipping benchmarks show that \method{} consistently outperforms representative baselines. Visualization analyses further show that \method{} provides interpretable reliability and modality-evidence diagnostics.
}



\maketitle
\section{Introduction}\label{sec:introduction}

Multimodal molecular representation learning aims to learn predictive molecular embeddings from multiple views of the same compound~\cite{edwards2022translation,liu2023multi,liu2023molca}. Typically, each molecule is represented by a structured molecular graph that characterizes atoms, chemical bonds, and topological connectivity~\cite{duvenaud2015convolutional,ma2022cross,wang2024chain}, together with associated textual descriptions, SMILES-based language representations~\cite{smiles,ross2022large,pei2023biot5}, or descriptor summaries that capture complementary semantic, physicochemical, and pharmacological information. Because these views encode different aspects of molecular behavior, integrating them enables more comprehensive molecular representations and has become an effective strategy for bioactivity prediction, toxicity assessment, physicochemical property modeling, functional annotation, and downstream experimental prioritization~\cite{pei2024biot5+,zheng2025large,boldini2024machine,wang2026sgac}.

\begin{figure}[t]
    \centering
    \includegraphics[width=0.98\linewidth]{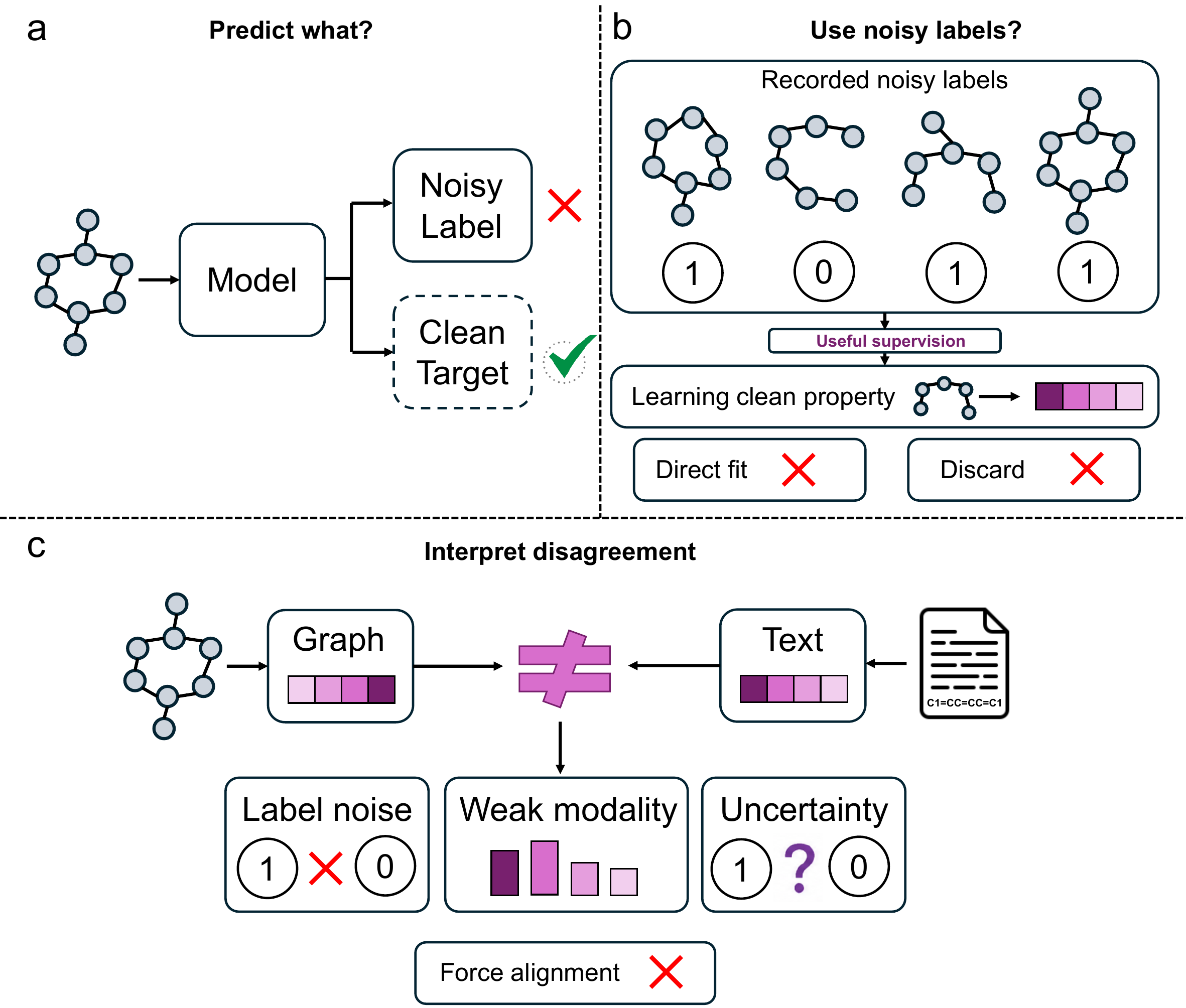}
     \vspace{-0.3cm}
    \caption{
    Overview of the three challenges studied in this work.
    \textbf{(a)} The prediction target should be separated from potentially noisy recorded labels.
    \textbf{(b)} Recorded labels should still provide useful supervision for learning clean molecular properties.
    \textbf{(c)} Graph--text disagreement may indicate label noise, weak modality evidence, or uncertainty, rather than a simple alignment error.
    }
    \vspace{-1.0cm}
    \label{fig:motivation}
\end{figure}

Molecular representation learning has evolved from manually designed chemical encodings to deep graph-based and multimodal paradigms~\cite{mcgibbon2024intuition,wang2025dusego}. Early approaches used molecular fingerprints, physicochemical descriptors, and SMILES strings as compact representations for similarity search and property prediction~\cite{rogers2010extended,deng2023systematic,wang2022advanced,wang2026riemannian}. With the development of deep learning, SMILES strings and molecular texts have been modeled by recurrent neural networks, convolutional networks, Transformers, and chemical language models, enabling the extraction of syntactic patterns, functional groups, and chemical semantics from sequential molecular descriptions~\cite{ross2022large,edwards2022translation,pei2023biot5}. However, sequence-based representations provide only an indirect description of molecular topology~\cite{yoshikai2024difficulty,sadeghi2024can,wang2026usbd}. Graph neural networks address this limitation by representing molecules as atom--bond graphs and propagating information over chemical neighborhoods, thereby capturing local substructures, long-range connectivity, and topology-dependent molecular patterns~\cite{ma2022cross,wang2024chain,zhao2024molecular}. More recently, multimodal molecular learning has sought to integrate graph- and text-derived information to obtain richer molecular representations~\cite{wu2023molecular,rollins2024molprop, zhang2024mvmrl}. Existing methods commonly combine modalities through cross-modal attention, contrastive alignment, shared embedding spaces, or molecule--language pretraining objectives~\cite{edwards2022translation,liu2023multi,liu2023molca}. These strategies improve representation capacity by exploiting the complementarity between structural and semantic molecular views~\cite{liu2023multi,liu2023molca,fang2024mol}. Despite this progress, molecular property labels are not always clean supervision. Recorded labels may be affected by measurement variability, assay interference, thresholding decisions, inconsistent experimental protocols, conflicting annotations, database curation, or weak automatic labeling pipelines~\cite{buterez2023mf,boldini2024machine,deng2023systematic}. When such labels are directly treated as ground truth, predictive models may memorize corrupted observations and learn misleading molecular evidence, leading to degraded generalization on more reliable evaluation data~\cite{wei2023fine,nguyen2024noisy,lin2024learning}. Although noisy-label learning has been widely studied, existing methods mainly aim to reduce label memorization in unimodal classification through robust losses, co-training, sample reweighting, semi-supervised label refinement, or meta reweighting~\cite{wei2023fine,lin2024learning}. These approaches typically regard corrupted labels as sample-level training noise, but they do not explicitly distinguish latent molecular properties from recorded noisy labels, nor do they address how complementary molecular views should be used when supervision is unreliable. This gap becomes especially important in multimodal molecular representation learning. Most multimodal objectives emphasize stronger graph--text fusion or alignment, implicitly assuming that the recorded label provides trustworthy supervision~\cite{edwards2022translation,liu2023multi,liu2023molca,wang2026nested}. Under noisy supervision, this assumption can be problematic: a corrupted label may be memorized by the predictor and further propagated across graph and text representations through the learned alignment. Moreover, graph--text disagreement may reflect label corruption, insufficient evidence in one modality, or genuine uncertainty about the molecular property, rather than a simple alignment error.

In this paper, we study noisy-label multimodal molecular representation learning and aim to develop a principled framework. As shown in Figure~\ref{fig:motivation}, this setting raises three key challenges. First, \textit{what should the model predict when recorded labels may be unreliable?} Molecular datasets usually provide only recorded labels, whereas the underlying molecular property of interest is not directly observed~\cite{boldini2024machine,deng2023systematic,wang2024degree}. Directly treating recorded labels as target variables may cause the model to absorb experimental variation, curation errors, or annotation noise as molecular evidence~\cite{wei2023fine,nguyen2024noisy,lin2024learning}. Second, \textit{how can noisy recorded labels still provide useful supervision?} Although recorded labels may be corrupted, they remain the main source of supervision. A noise-aware formulation should therefore connect the clean-property posterior to the recorded-label distribution, rather than either fitting recorded labels directly or discarding them~\cite{liu2023identifiability,liao2025instance,nguyen2024noisy,wang2026brain}. Third, \textit{how should graph--text disagreement be interpreted under noisy supervision?} In multimodal molecular learning, disagreement between graph and text views may reflect label corruption, insufficient evidence in one modality, or genuine uncertainty about the molecular property~\cite{edwards2022translation,liu2023multi}. Indiscriminately enforcing graph--text agreement may suppress useful modality-specific evidence and propagate label-induced errors~\cite{liu2023multi,liu2023molca,wei2023fine}.

To address these challenges, we propose \method{}, a noise-aware multimodal framework for learning molecular representations from noisy labels. Rather than treating recorded labels as clean targets, \method{} explicitly separates clean molecular property inference from recorded-label observation. Specifically, graph and text views are first encoded into modality-specific representations and then formulated as residual natural-parameter evidence for a latent categorical clean-property distribution. To use recorded labels without directly fitting them as clean supervision, \method{} introduces a categorical label-observation channel that maps the clean-property posterior to the recorded-label distribution. This probabilistic formulation links latent molecular properties to noisy supervision and naturally derives posterior label reliability from the model. To handle graph--text disagreement under unreliable supervision, \method{} regularizes high-confidence contradictory evidence between modalities while preserving modality-specific information. In addition, a perturbation-consistent clean-posterior regularizer improves stability under label-preserving molecular perturbations. To validate the effectiveness of \method{}, we conduct experiments on naturally noisy molecular benchmarks~\cite{buterez2023mf} and controlled label-flipping benchmarks~\cite{wu2018moleculenet}, showing that \method{} achieves state-of-the-art performance over representative graph-only, multimodal, and noisy-label learning baselines while providing interpretable posterior reliability and modality-specific molecular evidence.

Our contributions are summarized as follows: (1) We formulate noisy-label multimodal molecular representation learning around three challenges: separating latent molecular properties from recorded labels, using recorded labels as supervision through a label-observation channel, and interpreting graph--text disagreement under unreliable supervision. (2) We propose \method{}, a noise-aware framework that composes graph and text views as residual natural-parameter evidence for clean-property prediction and connects this prediction to recorded labels through a categorical label-observation channel. (3) We conduct experiments on naturally noisy molecular benchmarks and controlled label-flipping benchmarks, demonstrating state-of-the-art performance over representative graph-based, multimodal, and noisy-label learning baselines, together with interpretable posterior reliability and modality-specific molecular evidence.
\section{Materials and methods}

\subsection{Preliminary}

Given a molecule represented by a molecular graph $G=(V,E,X)$, where $V$ is the set of atoms, $E$ is the set of chemical bonds, and $X$ is the atom-feature matrix, together with a text-derived molecular view $T$, we study noisy-label multimodal molecular property prediction. The recorded label $\tilde y\in\mathcal Y$ may be affected by experimental variation, database curation, or weak annotation, and is therefore treated as a noisy observation of the latent clean molecular property label $y\in\mathcal Y$. Given a noisy multimodal training set $\mathcal D=\{(G_i,T_i,\tilde y_i)\}_{i=1}^{N}$, where $G_i$ and $T_i$ are the graph and text views of molecule $i$, our goal is to learn a clean molecular property posterior
\begin{equation}
\mathbf p_i
=
p_{\theta}
\left(
y_i\mid G_i,T_i
\right)
\in
\Delta^{C-1}.
\label{eq:clean-predictor}
\end{equation}
Here, $\mathcal Y=\{1,\ldots,C\}$ is the categorical label space, $\mathbf p_i=(p_{i,1},\ldots,p_{i,C})$ is the posterior used for inference, and $\Delta^{C-1}$ denotes the probability simplex over $\mathcal Y$, i.e. the set of non-negative $C$-dimensional vectors whose entries sum to one.

\subsection{Overview of \method{}}
\begin{figure*}
    \centering
    \includegraphics[width=1.0\linewidth]{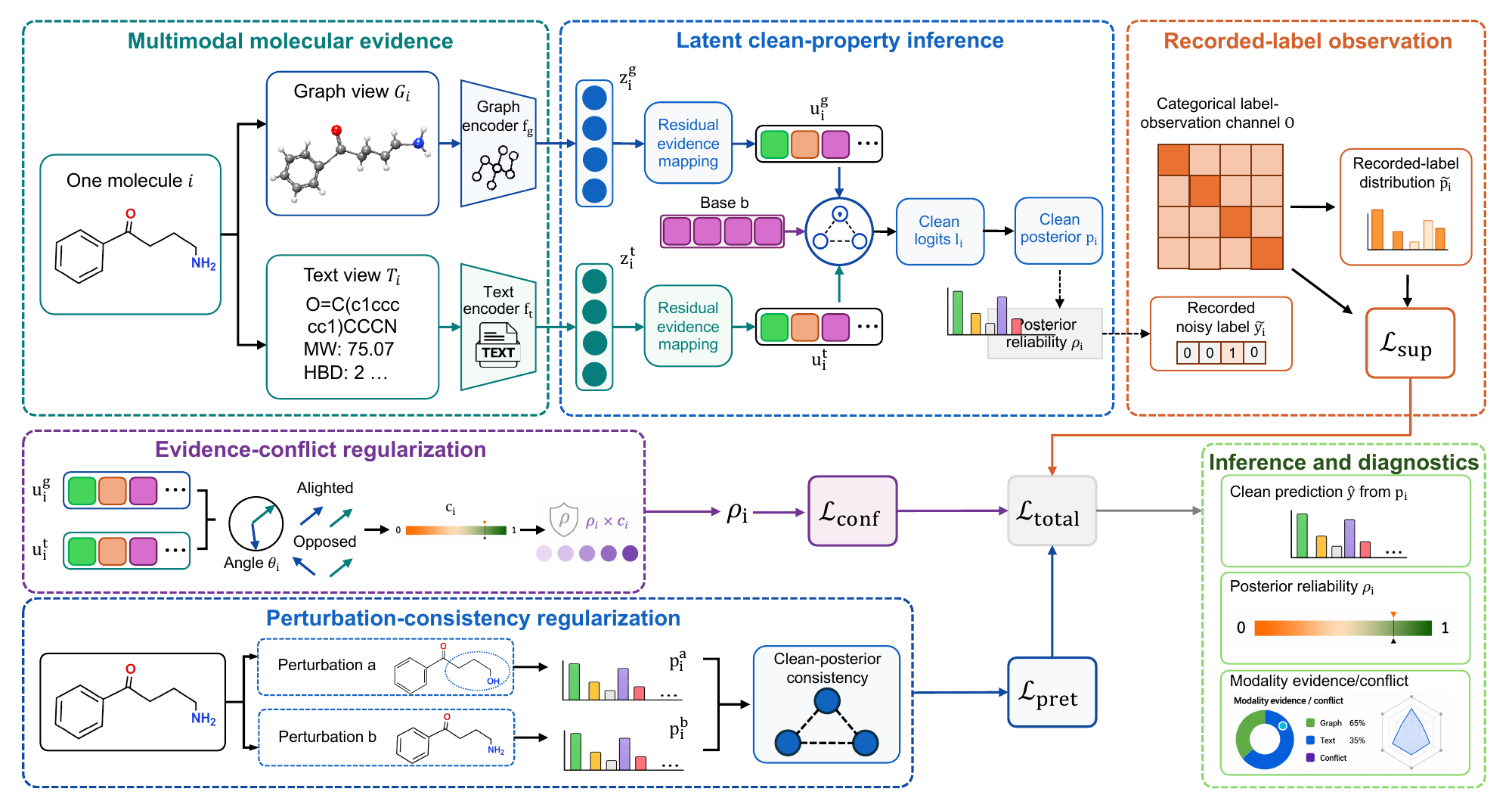}
    \caption{ Framework overview of \method{}. Graph and text views are composed into a latent clean-property posterior, which is connected to recorded noisy labels through a categorical label-observation channel. The framework also derives posterior reliability and regularizes evidence conflict and perturbation consistency. }
    \label{fig:overview}
\end{figure*}

As shown in Figure~\ref{fig:overview}, \method{} is a noise-aware framework for learning multimodal molecular representations from noisy labels. The framework contains four modules. The \textit{molecular evidence initialization} module encodes the molecular graph $G_i$ and the text-derived view $T_i$ into modality-specific representations. The \textit{residual natural-parameter evidence composition} module maps these representations to graph- and text-derived evidence and combines them into a latent clean categorical distribution. The \textit{categorical label-observation channel} links the clean posterior to the recorded-label distribution, allowing noisy labels to provide supervision without being treated as clean targets. The \textit{noise-aware learning objective} combines recorded-label likelihood with clean-evidence regularization to reduce contradictory graph--text evidence and improve perturbation-consistent clean posterior prediction.

\subsection{Molecular evidence initialization}

Given the graph and text-derived views of molecule $i$, \method{} first encodes them into modality-specific molecular representations. The graph view $G_i$ is processed by a graph encoder $f_g$, which can be instantiated by common message-passing architectures such as GCN, GAT, or GIN. The text-derived view \(T_i\) is encoded by a text-side encoder \(f_t\), which can be implemented using a pretrained molecular or biomedical language model, or a lightweight encoder over precomputed molecular text embeddings. The two representations are defined as
\begin{equation}
\mathbf z_i^g=f_g(G_i)\in\mathbb R^d,
\qquad
\mathbf z_i^t=f_t(T_i)\in\mathbb R^d .
\label{eq:modality-rep}
\end{equation}

The graph representation $\mathbf z_i^g$ summarizes structural information from the atom--bond topology, including local chemical neighborhoods and global connectivity patterns. The text representation $\mathbf z_i^t$ captures complementary semantic, physicochemical, or pharmacological information from molecular descriptions or text-derived embeddings. These two modality-specific representations serve as the initial molecular evidence for clean-property prediction.

\subsection{Residual natural-parameter evidence composition}

A common multimodal strategy is to concatenate graph and text representations, or to produce modality-specific predictions and combine them with a fusion module. Such designs can be fragile under noisy supervision: if the recorded label is corrupted, probability-level fusion or forced graph--text alignment may propagate label-induced errors across modalities. They also make it difficult to determine whether a prediction is supported by structural evidence, text-derived evidence, or both.

To preserve modality-specific evidence while forming a single clean-property predictor, \method{} composes graph and text information in the natural-parameter space of a categorical distribution. For a categorical variable with $C$ classes, the logit vector is a natural parameter and is identifiable up to an additive constant. In this space, additive residuals correspond to additive changes in class-relative evidence. We fix the arbitrary offset with the centering operator
\begin{equation}
\mathcal C(\mathbf v)
=
\mathbf v
-
\frac{1}{C}
\left(
\mathbf 1_C^{\top}\mathbf v
\right)
\mathbf 1_C,
\qquad
\mathbf v\in\mathbb R^C,
\label{eq:centering-map}
\end{equation}
where $\mathbf 1_C$ is the all-one vector. The operator removes the common additive component and keeps only class-relative evidence.

Given $\mathbf z_i^g$ and $\mathbf z_i^t$, two learnable evidence functions produce class-wise residual natural parameters:
\begin{equation}
\mathbf u_i^g
=
\mathcal C
\left(
\phi_g(\mathbf z_i^g)
\right),
\qquad
\mathbf u_i^t
=
\mathcal C
\left(
\phi_t(\mathbf z_i^t)
\right),
\label{eq:natural-evidence}
\end{equation}
where $\phi_g:\mathbb R^d\rightarrow\mathbb R^C$ and $\phi_t:\mathbb R^d\rightarrow\mathbb R^C$ are graph and text evidence functions, respectively. The component $u^g_{i,c}$ represents graph-derived residual evidence for class $c$, and $u^t_{i,c}$ represents text-derived residual evidence for the same class. Positive residual evidence increases the relative support for a class, whereas values close to zero indicate weak modality-specific evidence.

We further introduce a base natural-parameter vector $\mathbf b\in\mathbb R^C$ to represent class-level prior tendency. The clean-property logit is defined by additive evidence composition:
\begin{equation}
\boldsymbol\ell_i
=
\mathcal C
\left(
\mathbf b
+
\mathbf u_i^g
+
\mathbf u_i^t
\right).
\label{eq:clean-logit}
\end{equation}
The clean-property posterior is obtained by normalizing the clean natural parameters:
\begin{equation}
\mathbf p_i
=
\operatorname{softmax}
\left(
\boldsymbol\ell_i
\right),
\qquad
p_{i,c}
=
p_{\theta}
\left(
y_i=c\mid G_i,T_i
\right).
\label{eq:clean-softmax}
\end{equation}
Equations~\eqref{eq:natural-evidence}--\eqref{eq:clean-softmax} define a single latent clean-property distribution for molecule $i$. Instead of training independent graph, text, and fusion classifiers, the two modalities contribute residual evidence to the same categorical distribution. The base vector $\mathbf b$ captures class-level tendency, while $\mathbf u_i^g$ and $\mathbf u_i^t$ describe molecule-specific evidence from the graph and text views.

This explicit decomposition also supports modality-level interpretation. We quantify the relative contribution of graph-derived evidence by
\begin{equation}
M_i^g
=
\frac{
\left\|
\mathbf u_i^g
\right\|_2
}{
\left\|
\mathbf u_i^g
\right\|_2
+
\left\|
\mathbf u_i^t
\right\|_2
+
\varepsilon
},
\qquad
M_i^t
=
1-M_i^g,
\label{eq:modality-magnitude}
\end{equation}
where $\varepsilon>0$ is a small constant for numerical stability. A larger $M_i^g$ indicates stronger graph-derived evidence, whereas a larger $M_i^t$ indicates stronger text-derived evidence. These quantities are used for interpretation and diagnostic analysis, not as an additional fusion mechanism.

\subsection{Categorical label-observation channel}

The clean posterior $\mathbf p_i$ represents the predicted latent molecular property. However, the label available for training is the recorded label $\tilde y_i$, which may be affected by measurement variability, thresholding, database curation, or weak annotation. Directly optimizing the clean posterior against $\tilde y_i$ would implicitly treat the recorded label as the clean target. To avoid this assumption, \method{} introduces a categorical label-observation channel that links latent clean classes to recorded labels.

Let $\mathbf O\in[0,1]^{C\times C}$ denote the label-observation matrix:
\begin{equation}
O_{ac}
=
p_{\theta}
\left(
\tilde y_i=a
\mid
y_i=c
\right),
\qquad
a,c\in\mathcal Y .
\label{eq:observation-def}
\end{equation}
Here, $c$ indexes the latent clean class and $a$ indexes the recorded class. Each column of $\mathbf O$ is a categorical distribution over recorded labels conditioned on a clean class, satisfying $\sum_{a=1}^{C}O_{ac}=1$ for each $c$.

Since clean labels are not observed during training, an unconstrained channel may arbitrarily exchange latent clean classes and recorded labels. We therefore use a diagonal-dominant parameterization to stabilize this mapping while still allowing nonzero off-diagonal label transitions. Specifically, let
\begin{equation}
\eta_{ac}
=
\begin{cases}
0, & a=c,\\
-\zeta_{ac}, & a\neq c,
\end{cases}
\qquad
\zeta_{ac}>0,
\label{eq:observation-logit}
\end{equation}
where $\zeta_{ac}$ is a positive offset for the transition from clean class $c$ to recorded class $a$. The observation probabilities are obtained by a column-wise softmax:
\begin{equation}
O_{ac}
=
\frac{
\exp(\eta_{ac})
}{
\sum_{b=1}^{C}\exp(\eta_{bc})
}.
\label{eq:observation-prob}
\end{equation}
This parameterization assigns the largest logit in each column to the diagonal entry while retaining nonzero probabilities for class confusions. The channel is therefore a constrained probabilistic link between latent molecular properties and recorded labels, rather than an exact recovery of the underlying experimental or curation error process.

Given the clean-property posterior $\mathbf p_i$, the distribution over recorded labels is obtained by marginalizing the latent clean class through the label-observation channel:
\begin{equation}
\tilde{\mathbf p}_i
=
\left(
p_{\theta}
\left(
\tilde y_i=a
\mid
G_i,T_i
\right)
\right)_{a=1}^{C}
=
\mathbf O\mathbf p_i .
\label{eq:noisy-p-vector}
\end{equation}
Equivalently, for each recorded class $a\in\mathcal Y$,
\begin{equation}
\tilde p_{i,a}
=
\sum_{c=1}^{C}
O_{ac}p_{i,c}.
\label{eq:noisy-p-component}
\end{equation}
Thus, $\mathbf p_i$ is used for clean-property inference, whereas $\tilde{\mathbf p}_i$ is used to evaluate the likelihood of the recorded label.

The same channel also provides a posterior reliability score for the recorded label. Let $a_i=\tilde y_i$ denote the recorded class of molecule $i$. We define
\begin{equation}
\rho_i
=
p_{\theta}
\left(
y_i=a_i
\mid
\tilde y_i=a_i,
G_i,T_i
\right).
\label{eq:reliability-def}
\end{equation}
By Bayes' rule,
\begin{equation}
\rho_i
=
\frac{
O_{a_i a_i}p_{i,a_i}
}{
\sum_{c=1}^{C}
O_{a_i c}p_{i,c}
}
=
\frac{
O_{a_i a_i}p_{i,a_i}
}{
\tilde p_{i,a_i}
}.
\label{eq:reliability}
\end{equation}
Here, $\rho_i$ is the posterior probability that the recorded label coincides with the latent clean class under the learned clean posterior and label-observation channel. Thus, label reliability is obtained as a channel-derived posterior quantity rather than as an externally designed sample weight.

\subsection{Noise-aware learning objective}

The learning objective follows the separation between latent clean-property inference and noisy label observation. The recorded-label likelihood is evaluated after the clean posterior passes through the categorical label-observation channel. Two regularization terms are imposed on clean-property evidence: one controls contradictory graph--text evidence, and the other encourages perturbation-consistent clean posterior predictions.

\subsubsection{Recorded-label likelihood}

For molecule $i$, the clean posterior $\mathbf p_i$ is mapped to the recorded-label distribution $\tilde{\mathbf p}_i=\mathbf O\mathbf p_i$. The supervised likelihood is defined as the categorical negative log-likelihood of the recorded label:
\begin{equation}
\mathcal L_{\mathrm{sup}}
=
-
\frac{1}{N}
\sum_{i=1}^{N}
\log
\left(
\tilde p_{i,\tilde y_i}
+
\varepsilon
\right).
\label{eq:supervised-loss}
\end{equation}
This objective uses recorded labels through the label-observation channel, rather than directly treating them as clean targets.

\subsubsection{Posterior-weighted evidence-conflict regularization}

To reduce harmful cross-modal contradiction, we regularize the residual evidence from graph and text. We first define bounded evidence vectors
\begin{equation}
\mathbf s_i^g
=
\tanh
\left(
\mathbf u_i^g
\right),
\qquad
\mathbf s_i^t
=
\tanh
\left(
\mathbf u_i^t
\right),
\label{eq:bounded-evidence}
\end{equation}
where the operation is applied element-wise. Their inner product measures whether the two modalities support compatible class directions. The evidence-conflict score is
\begin{equation}
c_i
=
\frac{1}{C}
\left[
-
\left\langle
\mathbf s_i^g,
\mathbf s_i^t
\right\rangle
\right]_{+},
\qquad
[x]_{+}
=
\max(x,0).
\label{eq:conflict-score}
\end{equation}
The conflict regularizer is weighted by the posterior reliability $\rho_i$:
\begin{equation}
\mathcal L_{\mathrm{conf}}
=
\frac{1}{N}
\sum_{i=1}^{N}
\rho_i c_i .
\label{eq:conflict-loss}
\end{equation}
Thus, contradictory high-magnitude evidence is penalized more strongly when the recorded label is likely to match the latent clean class, while weak or compatible modality evidence contributes little to the penalty.

\subsubsection{Perturbation-consistent clean posterior regularization}

We further encourage the clean posterior to remain stable under label-preserving molecular perturbations. Let $\boldsymbol\ell_i^{(a)}$ and $\boldsymbol\ell_i^{(b)}$ be the clean logits obtained from two perturbed views of the same molecule. We define the centered posterior score as
\begin{equation}
\mathbf q
\left(
\boldsymbol\ell
\right)
=
\operatorname{softmax}
\left(
\boldsymbol\ell
\right)
-
\frac{1}{C}
\mathbf 1_C .
\label{eq:centered-posterior-score}
\end{equation}
The perturbation-consistency regularizer is
\begin{equation}
\mathcal L_{\mathrm{pert}}
=
\frac{1}{N}
\sum_{i=1}^{N}
\frac{1}{C}
\left\|
\mathbf q
\left(
\boldsymbol\ell_i^{(a)}
\right)
-
\mathbf q
\left(
\boldsymbol\ell_i^{(b)}
\right)
\right\|_2^2 .
\label{eq:perturbation-loss}
\end{equation}
Since the centered score is zero for a uniform categorical posterior and bounded across classes, this term stabilizes clean-property predictions without enforcing over-confident outputs.

\subsubsection{Unified training objective}

The final objective is
\begin{equation}
\mathcal L
=
\mathcal L_{\mathrm{sup}}
+
\beta
\mathcal L_{\mathrm{conf}}
+
\gamma
\mathcal L_{\mathrm{pert}},
\label{eq:total-loss}
\end{equation}
where $\beta\ge0$ and $\gamma\ge0$ control the strengths of evidence-conflict regularization and perturbation-consistent clean posterior regularization, respectively. 
\section{Experiments}

\subsection{Experimental setting}

\begin{figure*}[t]
    \centering
    \includegraphics[width=1.0\linewidth]{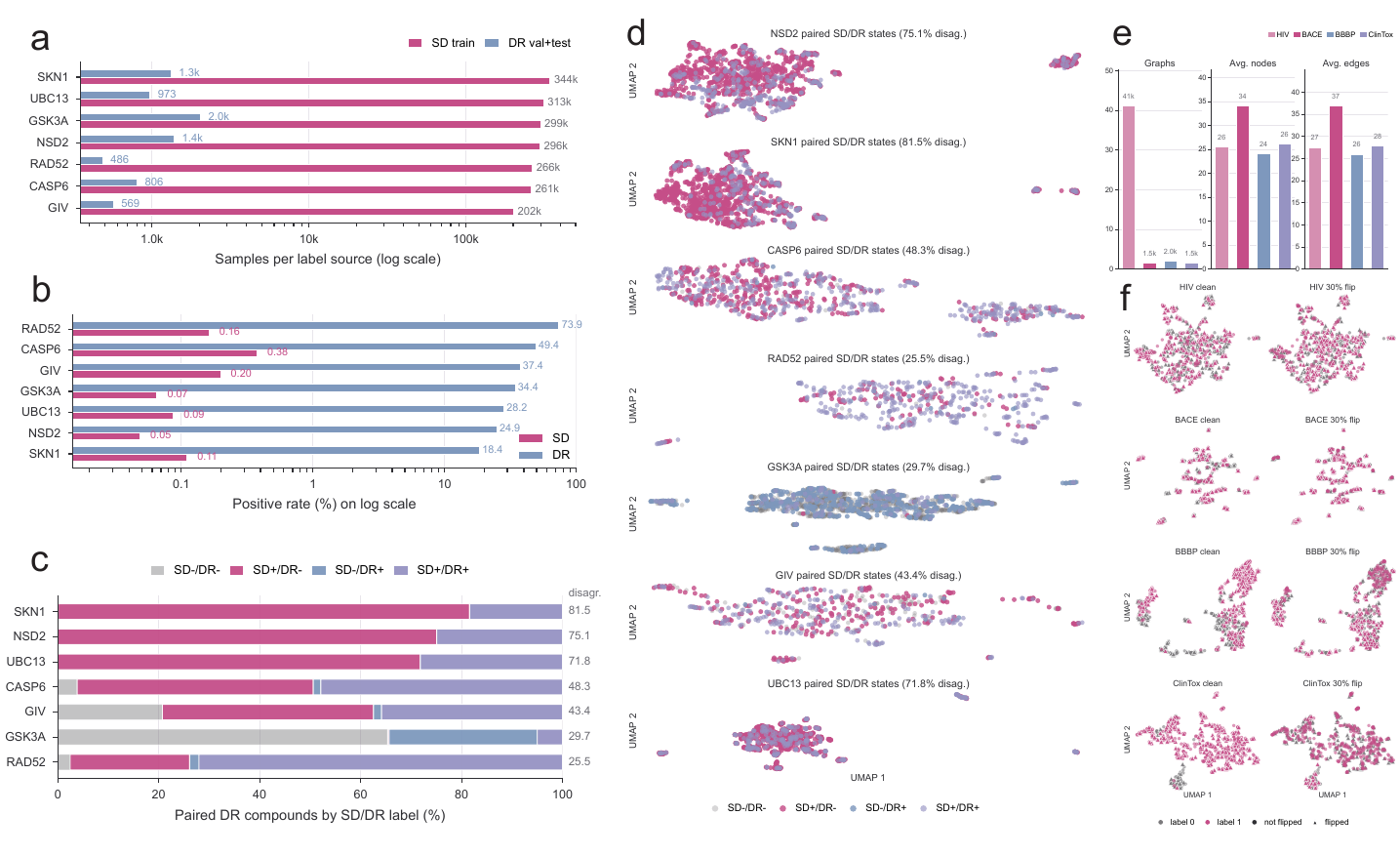}
    \caption{
    Dataset characteristics for natural and controlled noisy-label evaluation.
    \textbf{a--d}, Statistics of the naturally noisy \datasets{} benchmark, including the scale difference between single-dose (SD) screening labels and dose-response (DR) confirmatory labels, class imbalance, and SD--DR label disagreement.
    \textbf{e--f}, Controlled label-flipping protocol on MoleculeNet benchmarks, where training and validation labels are corrupted while test labels remain unchanged.
    }
    \label{fig:dataset_statistics}
\end{figure*}

\textbf{Datasets.}
We evaluate \method{} on two complementary groups of molecular property benchmarks: (1) The first group is \datasets{}, a naturally noisy molecular activity benchmark constructed from MF-PCBA~\cite{buterez2023mf}. MF-PCBA links large-scale primary screening measurements with confirmatory assays, making it suitable for studying how molecular predictors behave when recorded training labels are not fully reliable. In \datasets{}, single-dose (SD) screening labels are used as noisy training labels, while dose-response (DR) confirmatory labels are used as higher-confidence validation and test labels. This split reflects a realistic experimental scenario: SD assays provide broad but noisy coverage over many compounds, whereas DR assays are smaller but more reliable because activity is confirmed over multiple concentrations. As shown in Figure~\ref{fig:dataset_statistics}a--d, \datasets{} contains seven molecular activity prediction tasks: NSD2, SKN1, CASP6, RAD52, GSK3A, GIV, and UBC13. The SD training sets contain approximately 0.20--0.34 million molecules per task, whereas the DR validation and test sets contain hundreds to about one thousand molecules. The SD labels are extremely imbalanced, with positive rates below 0.4\%, while the DR positive rates are substantially higher. The paired SD--DR disagreement rates range from 25.51\% to 81.54\%. (2) The second group consists of four standard MoleculeNet benchmarks~\cite{wu2018moleculenet}: HIV, BACE, BBBP, and ClinTox. These datasets cover different molecular property prediction scenarios, including antiviral activity, enzyme inhibition, blood--brain barrier penetration, and clinical toxicity. Unlike \datasets{}, these benchmarks do not provide paired noisy and confirmatory labels. We therefore use them to construct controlled noisy-label settings with known corruption rates. For each dataset, molecules are split into five folds with a 3:1:1 train/validation/test allocation. Symmetric label-flip noise is injected into the training and validation labels, while the test labels remain unchanged. Unless otherwise specified, the label-flipping rate is fixed at 30\%. This controlled setting complements \datasets{} by evaluating robustness when the corruption process is known and the held-out test labels remain clean. The statistics of the above datasets are summarized in Table~\ref{tab:datasets} and ~\ref{tab:dataset_control}.

\noindent\textbf{Baselines.}
We compare \method{} with representative baselines from four categories. The first category includes supervised graph neural networks, namely GCN~\cite{GCN}, GAT~\cite{GAT}, and GIN~\cite{gin}. The second category includes molecular representation learning and pretraining methods, including GraphCL~\cite{GraphCL}, GROVE~\cite{GROVE}, SmiSGT~\cite{SmiSGT}, Uni-Mol~\cite{uni-mol}, and S-GCIB~\cite{s-gcib}. The third category includes multimodal molecular learning methods, including Tri-SGD~\cite{Tri-SGD}, MMSG~\cite{MMSG}, MDFCL~\cite{mdfcl}, and ProtoMol~\cite{protomol}. The fourth category includes robust graph or noisy-label learning methods, including OMG~\cite{omg}, RTGNN~\cite{rtgnn}, SPORT~\cite{sport}, and TFR~\cite{TFR}. More details of these baselines are provided in Appendix~\ref{app:baseline}.

\noindent\textbf{Evaluation metrics and protocol.}
The primary metric is the area under the receiver operating characteristic curve (ROC-AUC), denoted as AUC. Given prediction scores $\{s_i\}$ and binary labels $\{y_i\}$, let $\mathcal P=\{i:y_i=1\}$ and $\mathcal N=\{j:y_j=0\}$ denote the positive and negative sample sets. AUC can be written as
\begin{equation}
\mathrm{AUC}
=
\frac{1}{|\mathcal P||\mathcal N|}
\sum_{i\in\mathcal P}
\sum_{j\in\mathcal N}
\left[
\mathbb I(s_i>s_j)
+
\frac{1}{2}\mathbb I(s_i=s_j)
\right].
\end{equation}
AUC is threshold independent and measures whether positive samples are ranked above negative samples. Because several tasks are highly imbalanced, we also report the area under the precision--recall curve (AUPRC) and Matthews correlation coefficient (MCC). For a ranked prediction list, AUPRC is computed as
\begin{equation}
\mathrm{AUPRC}
=
\sum_{r=1}^{R}
\left(\mathrm{Recall}_{r}-\mathrm{Recall}_{r-1}\right)
\mathrm{Precision}_{r},
\end{equation}
where $r$ indexes operating points along the precision--recall curve. MCC is defined as
\begin{equation}
\mathrm{MCC}
=
\frac{\mathrm{TP}\cdot\mathrm{TN}-\mathrm{FP}\cdot\mathrm{FN}}
{
\sqrt{
(\mathrm{TP}+\mathrm{FP})
(\mathrm{TP}+\mathrm{FN})
(\mathrm{TN}+\mathrm{FP})
(\mathrm{TN}+\mathrm{FN})
}
},
\end{equation}
where TP, TN, FP, and FN denote true positives, true negatives, false positives, and false negatives, respectively.

\noindent\textbf{Implementation details.}
All methods are implemented in PyTorch~\footnote{https://pytorch.org/}. Molecular graphs are constructed from SMILES using RDKit~\footnote{https://www.rdkit.org/} with standard atom and bond features, and text-derived views are generated from rule-based molecular descriptions, encoded by BiomedBERT, and cached before training. No atom or bond deletion is applied in the default perturbation pipeline, so the molecular topology is preserved. For \method{}, the graph branch uses a GIN encoder~\cite{gin}, and the text branch uses a lightweight multilayer encoder. The hidden dimension is 256, dropout is 0.2, the learning rate is $3\times10^{-4}$, the weight decay is $10^{-5}$, and the batch size is 512. Models are optimized with Adam and selected by validation performance. The label-observation channel is initialized close to the identity mapping, and the perturbation-consistency term uses label-preserving stochastic views, including atom-feature masking and text-view dropout. The regularization weights $\beta$ and $\gamma$, together with channel-related hyperparameters, are selected from small validation grids. For \datasets{}, models are trained on noisy SD labels, selected on the DR validation split, and evaluated on the held-out DR test split. All experiments are repeated with five random seeds under the same settings.

\subsection{Performance comparison}
\begin{table*}[t]
\centering
\caption{Classification results (ROC-AUC $\pm$ Std \%) on MF-PCBA-Noisy7 and MoleculeNet benchmarks.}
\label{tab:results_noise}
\resizebox{\textwidth}{!}{
\begin{tabular}{l ccccccccccc}
\toprule
\multirow{2}{*}{\textbf{Method}}
& \multicolumn{7}{c}{\textbf{MF-PCBA-Noisy7}} 
& \multicolumn{4}{c}{\textbf{MoleculeNet}} \\
\cmidrule(lr){2-8} \cmidrule(lr){9-12}
& \textbf{NSD2}
& \textbf{SKN1}
& \textbf{CASP6}
& \textbf{RAD52}
& \textbf{GSK3A}
& \textbf{GIV}
& \textbf{UBC13}
& \textbf{HIV}
& \textbf{BACE}
& \textbf{BBBP}
& \textbf{ClinTox} \\
\midrule
GCN
& $76.1 \pm 0.8$ & $72.5 \pm 1.1$ & $67.9 \pm 1.4$ & $63.2 \pm 1.2$ & $56.4 \pm 0.9$ & $45.3 \pm 1.3$ & $43.1 \pm 1.0$
& $59.1 \pm 9.6$ & $66.5 \pm 5.6$ & $71.5 \pm 6.1$ & $56.7 \pm 7.2$ \\

GAT
& $74.6 \pm 1.2$ & $65.2 \pm 0.9$ & $68.1 \pm 1.5$ & $62.4 \pm 0.7$ & $57.8 \pm 1.1$ & $42.2 \pm 1.4$ & $41.6 \pm 0.8$
& $55.3 \pm 7.0$ & $52.6 \pm 3.4$ & $64.6 \pm 3.9$ & $50.7 \pm 4.9$ \\

GIN
& $71.9 \pm 0.9$ & $70.9 \pm 1.3$ & $66.3 \pm 1.0$ & $64.5 \pm 1.4$ & $55.5 \pm 0.8$ & $46.8 \pm 1.1$ & $44.5 \pm 1.2$
& $62.2 \pm 3.2$ & $59.1 \pm 10.8$ & $78.3 \pm 3.2$ & $55.4 \pm 8.0$ \\
\midrule
GraphCL
& $77.9 \pm 1.2$ & $76.7 \pm 0.9$ & $73.6 \pm 1.1$ & $70.9 \pm 0.8$ & $62.7 \pm 1.4$ & $57.4 \pm 1.3$ & $50.1 \pm 1.0$
& $61.6 \pm 4.3$ & $57.9 \pm 7.0$ & $75.5 \pm 2.5$ & $57.4 \pm 10.7$ \\

GROVE
& $82.4 \pm 0.7$ & $74.9 \pm 1.4$ & $73.7 \pm 1.2$ & $72.0 \pm 0.9$ & $65.1 \pm 1.5$ & $51.1 \pm 1.1$ & $50.4 \pm 0.8$
& $62.9 \pm 2.3$ & $66.4 \pm 6.4$ & $80.2 \pm 2.4$ & $56.3 \pm 9.5$ \\

SmiSGT
& $80.4 \pm 1.3$ & $77.1 \pm 0.8$ & $74.5 \pm 1.0$ & $65.6 \pm 1.4$ & $63.3 \pm 0.9$ & $54.2 \pm 1.2$ & $50.1 \pm 0.7$
& $58.9 \pm 5.8$ & $62.9 \pm 7.1$ & $77.2 \pm 2.2$ & $54.1 \pm 8.8$ \\

Uni-Mol
& $78.0 \pm 1.1$ & $75.6 \pm 1.5$ & $75.5 \pm 0.8$ & $67.7 \pm 1.3$ & $64.6 \pm 1.0$ & $46.3 \pm 0.9$ & $51.6 \pm 1.4$
& $56.4 \pm 6.7$ & $58.1 \pm 4.7$ & $73.7 \pm 1.5$ & $53.6 \pm 4.2$ \\

S-GCIB
& $84.2 \pm 0.9$ & $76.5 \pm 1.2$ & $74.1 \pm 0.7$ & $72.7 \pm 1.1$ & $62.2 \pm 1.3$ & $48.8 \pm 0.8$ & $50.3 \pm 1.0$
& $60.0 \pm 3.2$ & $66.4 \pm 7.5$ & $80.6 \pm 1.6$ & $60.6 \pm 9.2$ \\
\midrule
Tri-SGD
& $76.6 \pm 1.4$ & $73.1 \pm 1.0$ & $73.0 \pm 0.8$ & $59.9 \pm 1.2$ & $64.4 \pm 1.5$ & $44.4 \pm 0.9$ & $56.8 \pm 1.3$
& $62.5 \pm 3.2$ & $63.2 \pm 4.8$ & $78.7 \pm 2.8$ & $53.0 \pm 3.7$ \\

MMSG
& $75.8 \pm 0.7$ & $68.1 \pm 1.2$ & $72.6 \pm 1.3$ & $61.1 \pm 0.9$ & $63.6 \pm 1.4$ & $51.8 \pm 1.1$ & $49.2 \pm 0.8$
& $61.3 \pm 3.7$ & $61.1 \pm 4.0$ & $83.1 \pm 4.9$ & $74.2 \pm 5.8$ \\

MDFCL
& $74.9 \pm 1.1$ & $69.1 \pm 0.8$ & $71.5 \pm 1.0$ & $66.9 \pm 1.5$ & $59.4 \pm 0.7$ & $54.0 \pm 1.3$ & $51.5 \pm 1.4$
& $66.6 \pm 3.7$ & $68.4 \pm 8.6$ & $85.3 \pm 4.6$ & $79.2 \pm 10.8$ \\

Protomol
& $77.3 \pm 1.2$ & $73.8 \pm 1.1$ & $73.7 \pm 0.9$ & $60.7 \pm 1.0$ & $65.1 \pm 1.2$ & $45.2 \pm 0.8$ & $57.6 \pm 1.1$
& $63.4 \pm 2.9$ & $64.2 \pm 4.2$ & $84.5 \pm 2.7$ & $74.3 \pm 3.4$ \\
\midrule
OMG
& $82.1 \pm 0.9$ & $73.4 \pm 1.4$ & $65.8 \pm 1.1$ & $65.2 \pm 0.8$ & $61.6 \pm 0.3$ & $54.4 \pm 0.6$ & $52.6 \pm 1.0$
& $59.6 \pm 5.7$ & $63.4 \pm 5.0$ & $77.2 \pm 4.2$ & $58.0 \pm 7.6$ \\

SPORT
& $75.9 \pm 1.3$ & $77.5 \pm 0.7$ & $70.1 \pm 1.2$ & $60.3 \pm 1.4$ & $57.3 \pm 0.9$ & $53.8 \pm 0.8$ & $45.4 \pm 1.1$
& $57.6 \pm 4.5$ & $64.4 \pm 2.1$ & $73.9 \pm 3.8$ & $52.9 \pm 6.8$ \\

RTGNN
& $60.9 \pm 1.5$ & $54.1 \pm 0.8$ & $45.7 \pm 1.4$ & $66.5 \pm 1.1$ & $46.9 \pm 1.2$ & $53.5 \pm 0.9$ & $49.2 \pm 0.8$
& $61.4 \pm 4.0$ & $69.6 \pm 4.2$ & $76.5 \pm 6.8$ & $52.7 \pm 8.4$ \\

TFR
& $77.0 \pm 2.5$ & $70.5 \pm 7.1$ & $68.5 \pm 2.9$ & $70.7 \pm 1.3$ & $63.1 \pm 1.4$ & $55.7 \pm 2.9$ & $55.7 \pm 3.0$
& $62.5 \pm 3.0$ & $65.1 \pm 10.7$ & $75.0 \pm 3.7$ & $55.9 \pm 7.5$ \\

\midrule
\method{}
& $\textbf{84.6} \pm 0.7$ & $\textbf{78.9} \pm 0.6$ & $\textbf{76.2} \pm 0.4$ & $\textbf{74.8} \pm 1.1$ & $\textbf{66.5} \pm 0.6$ & $\textbf{60.2} \pm 1.0$ & $\textbf{57.9} \pm 0.6$
& $\textbf{67.7} \pm 3.0$ & $\textbf{74.0} \pm 1.7$ & $\textbf{89.9} \pm 2.2$ & $\textbf{82.2} \pm 7.8$ \\
\bottomrule
\end{tabular}
}
\vspace{-0.4cm}
\end{table*}

\begin{figure*}
    \centering
    \includegraphics[width=1.0\linewidth]{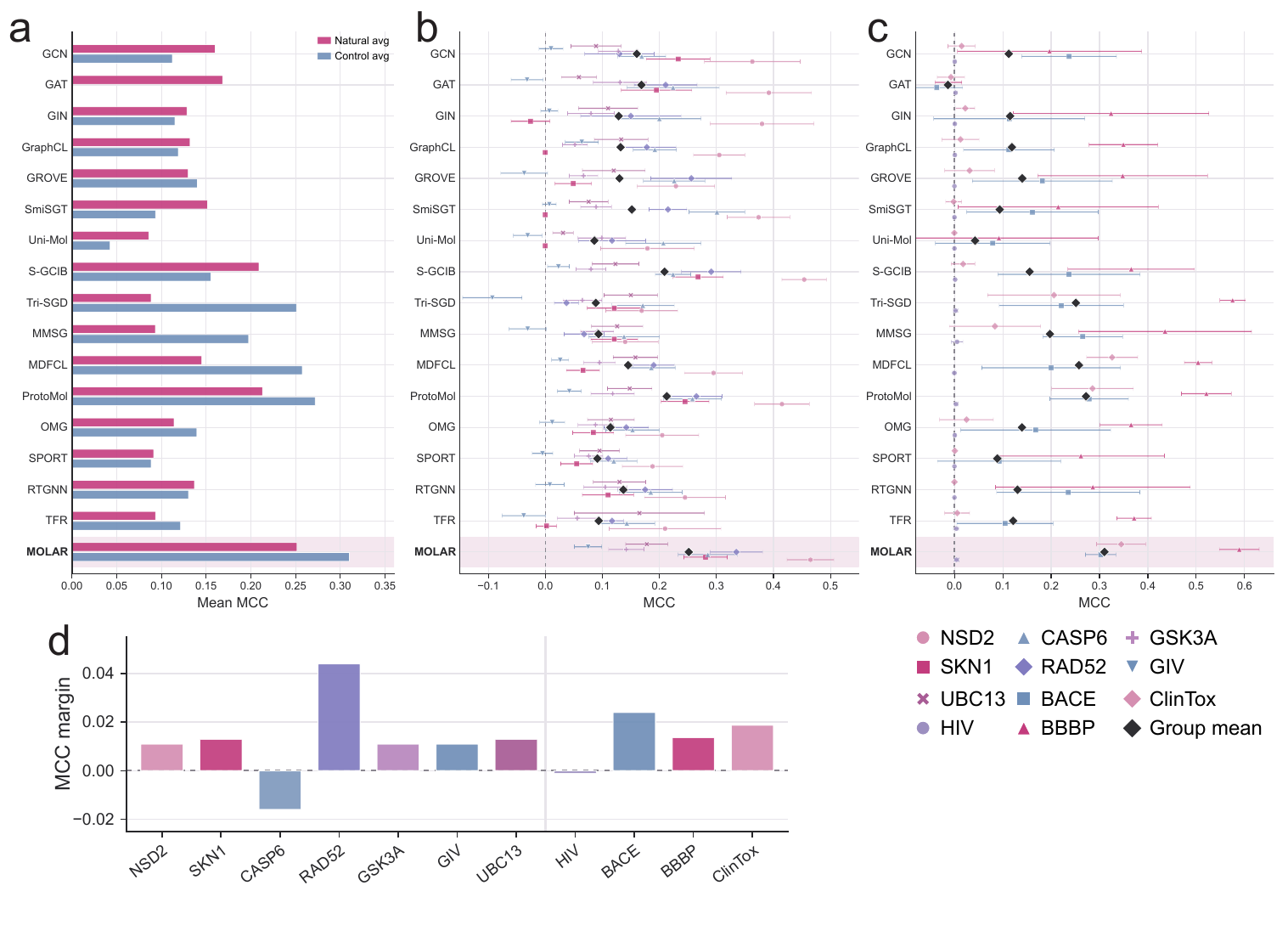}
    \caption{
    MCC-based comparison on natural-noise and control-noise tasks.
    (a) Average MCC over natural-noise and control-noise tasks for each method.
    (b,c) Per-task MCC on natural-noise MF-PCBA tasks and control-noise MoleculeNet tasks, respectively. Points denote mean MCC and horizontal bars denote standard deviation across runs; black diamonds indicate the group-wise mean for each method. The dashed vertical line marks zero MCC, and the shaded row highlights \method{}.
    (d) Dataset-wise margin between \method{} and the strongest baseline, computed as $\mathrm{MCC}_{\method{}}-\max_{\mathrm{baseline}}\mathrm{MCC}$.
    }
    \label{fig:mcc}
\end{figure*}

\begin{figure*}
    \centering
    \includegraphics[width=1.0\linewidth]{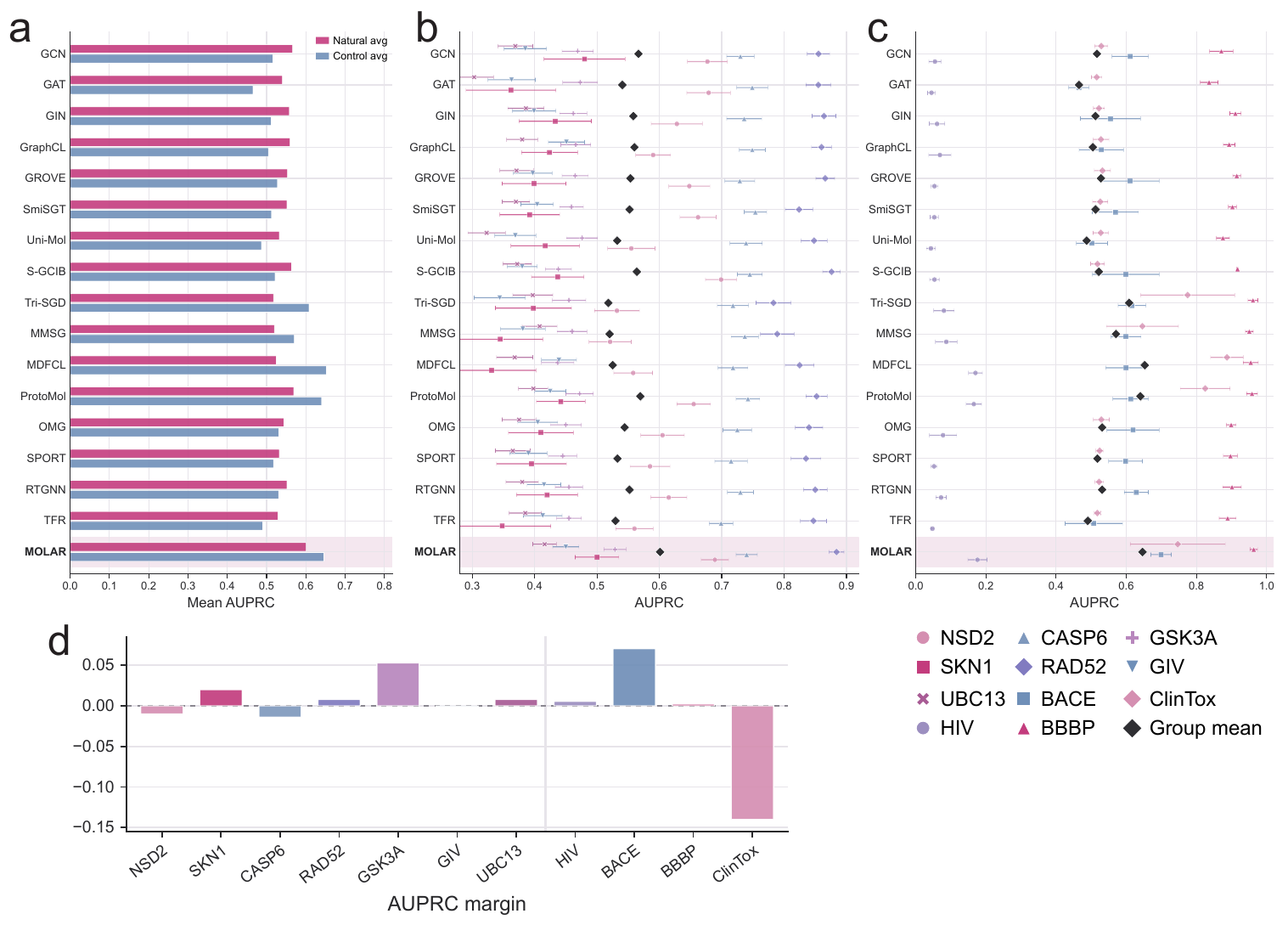}
    \caption{
    AUPRC-based comparison on natural-noise and control-noise tasks.
    (a) Average AUPRC over natural-noise and control-noise tasks for each method.
    (b,c) Per-task AUPRC on natural-noise MF-PCBA tasks and control-noise MoleculeNet tasks, respectively. Points denote mean AUPRC and horizontal bars denote standard deviation across runs; black diamonds indicate the group-wise mean for each method. The shaded row highlights \method{}.
    (d) Dataset-wise margin between \method{} and the strongest baseline, computed as $\mathrm{AUPRC}_{\method{}}-\max_{\mathrm{baseline}}\mathrm{AUPRC}$.
    }
    \label{fig:auprc}
\end{figure*}

Table~\ref{tab:results_noise}, Figure~\ref{fig:mcc} and \ref{fig:auprc} report the performance comparisons on MF-PCBA-Noisy7 and MoleculeNet benchmarks. Overall, \method{} achieves the best performance across the two benchmark groups, with consistent improvements over graph-based predictors, multimodal representation learning methods, and noisy-label learning baselines. On \datasets{}, the improvement is more pronounced for challenging tasks with severe label inconsistency, whereas the gains are relatively moderate on tasks where existing molecular representation learners are already competitive. On the controlled label-flipping benchmarks, \method{} also maintains a clear advantage under synthetic corruption. These trends suggest that the proposed framework is particularly beneficial when recorded labels are unreliable: by separating clean-property inference from noisy label observation and by modeling graph--text evidence explicitly, \method{} reduces the effect of corrupted supervision while preserving the complementary information provided by the two molecular views.

\begin{table}[t]
\centering
\caption{Aggregate comparison across all reported tasks. \textbf{Bold} results indicate the best performance.}
\label{tab:aggregate_results}
\resizebox{\columnwidth}{!}{
\begin{tabular}{lcccccc}
\toprule
\textbf{Method} 
& \textbf{Natural Avg.} 
& \textbf{Control Avg.} 
& \textbf{Overall Avg.} 
& \textbf{Avg. Rank} 
& \textbf{$p$-value} 
& \textbf{W/T/L} \\
\midrule
GCN      & 60.64 & 63.45 & 61.66 & 12.18 & $3.51{\times}10^{-5}$ & 10/1/0 \\
GAT      & 58.84 & 55.80 & 57.74 & 15.64 & $2.67{\times}10^{-5}$ & 11/0/0 \\
GIN      & 60.06 & 63.75 & 61.40 & 12.45 & $1.79{\times}10^{-5}$ & 10/1/0 \\
\midrule
GraphCL  & 67.04 & 63.10 & 65.61 & 7.86  & $3.36{\times}10^{-3}$ & 10/1/0 \\
GROVE    & 67.09 & 66.45 & 66.85 & 5.77  & $6.91{\times}10^{-3}$ & 8/3/0 \\
SmiSGT   & 66.46 & 63.28 & 65.30 & 8.18  & $3.30{\times}10^{-3}$ & 10/1/0 \\
Uni-Mol  & 65.61 & 60.45 & 63.74 & 9.27  & $2.05{\times}10^{-3}$ & 10/1/0 \\
S-GCIB   & 66.97 & 66.90 & 66.95 & 6.32  & $3.22{\times}10^{-3}$ & 8/3/0 \\
\midrule
Tri-SGD  & 64.03 & 64.35 & 64.15 & 9.68  & $2.60{\times}10^{-3}$ & 8/3/0 \\
MMSG     & 63.17 & 69.93 & 65.63 & 9.95  & $1.03{\times}10^{-5}$ & 8/3/0 \\
MDFCL    & 63.90 & 74.88 & 67.89 & 7.18  & $1.98{\times}10^{-5}$ & 7/4/0 \\
ProtoMol & 64.77 & 71.60 & 67.25 & 6.73  & $1.01{\times}10^{-3}$ & 7/4/0 \\
\midrule
OMG      & 65.01 & 64.55 & 64.85 & 8.77  & $4.59{\times}10^{-4}$ & 10/1/0 \\
SPORT    & 62.90 & 62.20 & 62.65 & 11.64 & $4.22{\times}10^{-4}$ & 11/0/0 \\
RTGNN    & 53.83 & 65.05 & 57.91 & 12.23 & $3.16{\times}10^{-4}$ & 9/2/0 \\
TFR      & 65.89 & 64.63 & 65.43 & 8.14  & $2.17{\times}10^{-3}$ & 8/3/0 \\
\midrule
\method{} & \textbf{71.30} & \textbf{78.45} & \textbf{73.90} & \textbf{1.00} & -- & -- \\
\bottomrule
\end{tabular}
}
\vspace{-0.4cm}
\end{table}

We further summarize the aggregate comparison in Table~\ref{tab:aggregate_results}. 
Natural Avg. denotes the mean ROC-AUC over the seven MF-PCBA-Noisy7 tasks, Control Avg. denotes the mean ROC-AUC over the four MoleculeNet benchmarks, and Overall Avg. is computed across all eleven tasks. 
Avg. Rank denotes the average task-wise rank, where a smaller value indicates better overall ranking. 
The $p$-value is computed using a paired task-level $t$-test over the eleven task-wise mean ROC-AUC values. For W/T/L, \method{} is compared with each baseline on every task using the reported mean $\pm$ standard-deviation intervals; non-overlapping intervals in favor of \method{} are counted as wins, overlapping intervals as ties, and non-overlapping intervals in favor of the baseline as losses. 
From these results, it can be observed that \method{} achieves strong and consistent performance across both benchmark groups. These findings suggest that the performance gains are systematic and robust across diverse noisy-label settings, rather than being driven by a small number of favorable tasks.

\subsection{Robustness Analysis}

\begin{figure*}
    \centering
    \includegraphics[width=1.0\linewidth]{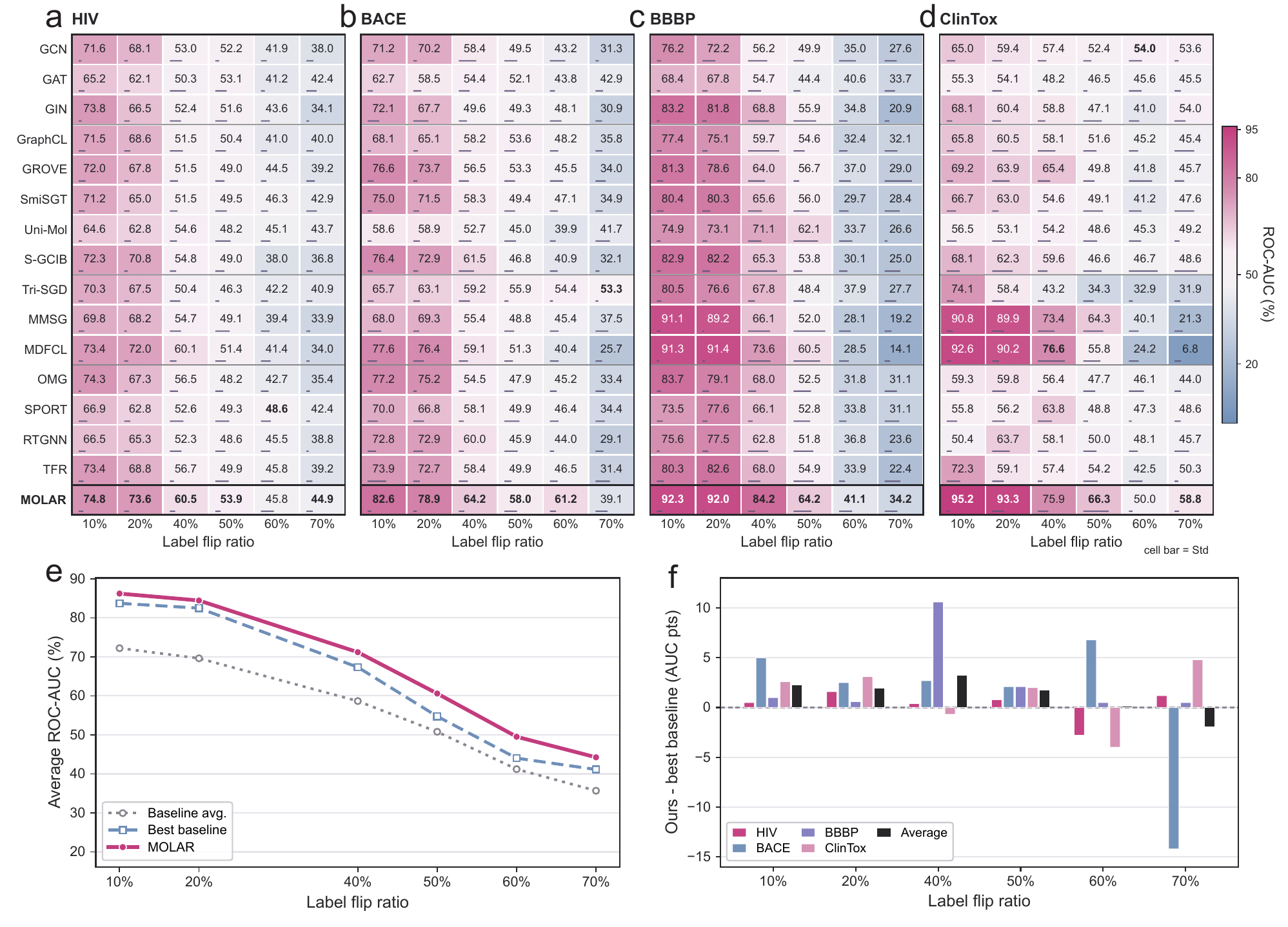}
    \caption{
    Control-noise robustness under increasing label-flip ratios.
    (a--d) ROC-AUC heatmaps for HIV, BACE, BBBP, and ClinTox across 10--70\% label flipping; cell values denote mean ROC-AUC and short bars indicate standard deviation.
    (e) Average ROC-AUC across the four datasets.
    (f) Margin between \method{} and the strongest baseline at each flip ratio.
    }
    \label{fig:sweep}
\end{figure*}

To evaluate robustness under different levels of label corruption, we further conduct controlled experiments with label-flipping rates ranging from 10\% to 70\% on MoleculeNet benchmarks. The results are reported in Figure~\ref{fig:sweep}. Across all settings, \method{} consistently ranks among the strongest methods and achieves the best overall performance under moderate corruption rates (10\%--40\%). As the noise rate increases, the performance of most baselines drops rapidly, whereas \method{} maintains comparatively stable prediction accuracy on several datasets, especially BBBP and ClinTox. This trend suggests that separating clean-property inference from noisy label observation enables the model to remain effective even when recorded labels become increasingly unreliable.

\subsection{Ablation study}
\begin{figure*}
    \centering
    \includegraphics[width=1.0\linewidth]{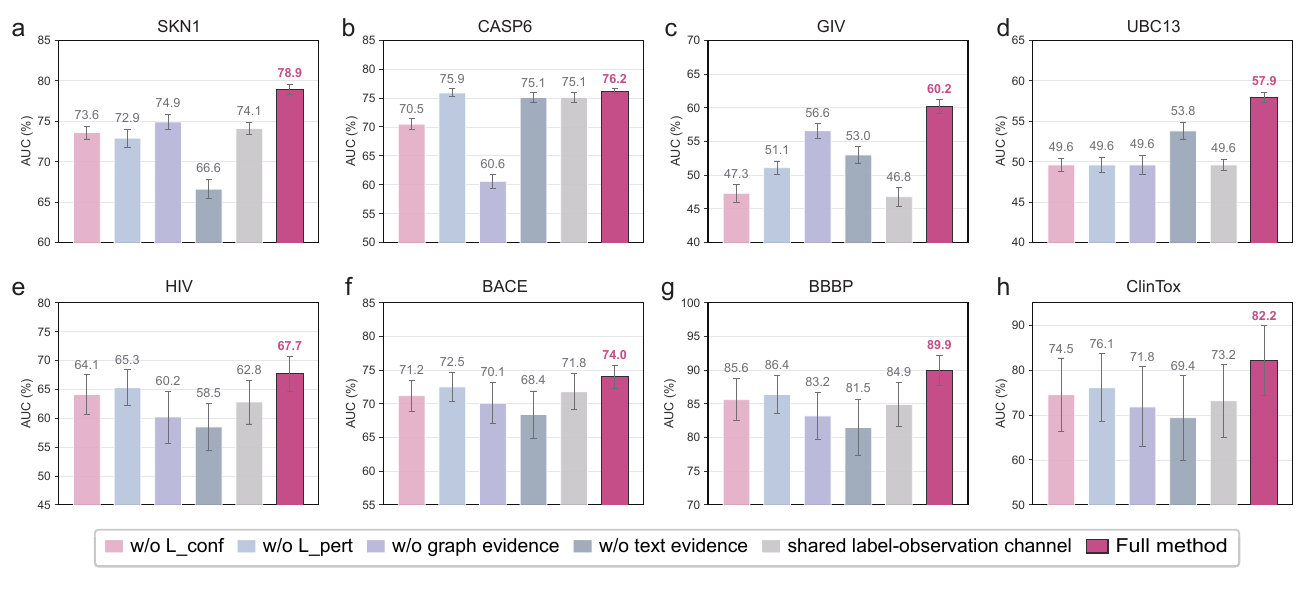}
    \caption{
    Ablation study on natural-noise and controlled-noise datasets. AUC scores are reported for different model variants, with error bars denoting standard deviations across runs.
    \textbf{a--d}, Results on natural-noise tasks.
    \textbf{e--h}, Results on controlled-noise tasks.
    }
    \label{fig:ablation}
\end{figure*}

To evaluate the contribution of each component in \method{}, we compare the full model with five variants: removing the evidence-conflict regularization term $\mathcal L_{\mathrm{conf}}$, removing the perturbation-consistency term $\mathcal L_{\mathrm{pert}}$, removing graph-derived evidence, removing text-derived evidence, and replacing the learned label-observation channel with a shared channel. As shown in Figure~\ref{fig:ablation}, the full model achieves the strongest performance across both natural-noise tasks and controlled-noise tasks. Removing $\mathcal L_{\mathrm{conf}}$ or $\mathcal L_{\mathrm{pert}}$ consistently weakens performance, indicating that both cross-modal evidence-conflict control and perturbation-consistent clean posterior learning are important for robustness under noisy supervision. The modality ablations further show that graph and text views provide complementary molecular evidence: removing either view leads to clear degradation, but the magnitude varies across tasks, suggesting that different molecular properties rely on different sources of information. The shared-channel variant also performs worse than the full model, supporting the need to explicitly learn a flexible label-observation channel rather than using an overly restricted noisy-label link. Overall, these results confirm that the performance gains of \method{} arise from the joint design of residual graph--text evidence, clean-posterior regularization, and noise-aware label observation.

\subsection{Hyperparameter analysis}
\begin{figure*}
    \centering
    \includegraphics[width=1.0\linewidth]{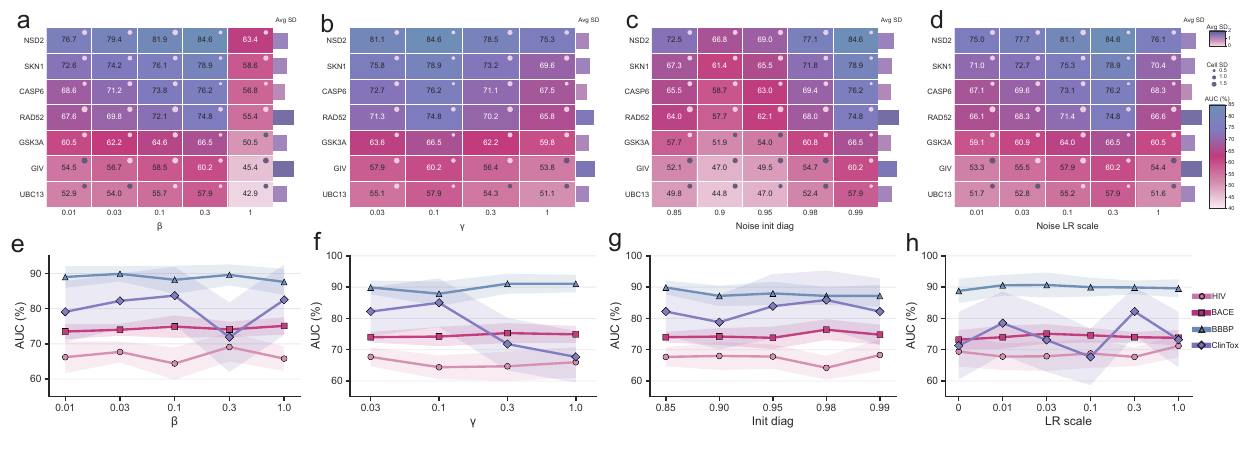}
    \caption{ Hyperparameter sensitivity of \method{}. \textbf{a--d}, Natural-noise AUC heatmaps under different values of $\beta$, $\gamma$, observation-channel diagonal initialization, and observation-channel learning-rate scale. \textbf{e--h}, Corresponding sensitivity curves on MoleculeNet datasets. Cell colors and line values indicate ROC-AUC.}
    \label{fig:sensitivity}
\end{figure*}

To examine the sensitivity of \method{} to its main hyperparameters, we vary the evidence-conflict weight $\beta$, the perturbation-consistency weight $\gamma$, the diagonal initialization of the label-observation channel, and the learning-rate scale for the channel parameters. Figure~\ref{fig:sensitivity} reports the results on both MF-PCBA-Noisy7 and MoleculeNet benchmarks. Overall, \method{} is stable across a reasonable range of settings. For $\beta$, performance generally improves from very small values to a moderate value, indicating that evidence-conflict regularization is beneficial, while an overly large $\beta$ can hurt performance by over-penalizing useful graph--text disagreement. For $\gamma$, moderate perturbation consistency gives the best results, whereas overly strong consistency tends to reduce performance on several natural-noise tasks. The label-observation channel parameters show a clearer trend: initializing the channel closer to the identity mapping and using a moderate channel learning-rate scale lead to better performance, especially on naturally noisy tasks where the relationship between latent properties and recorded labels is more uncertain. In contrast, the controlled label-flipping benchmarks show smoother sensitivity curves, suggesting that \method{} does not rely on a narrowly tuned configuration. These results highlight the importance of balancing clean-evidence regularization with a sufficiently flexible label-observation channel.

\subsection{Reliability diagnostics and molecular evidence visualization}

\begin{figure}
    \centering
    \includegraphics[width=0.98\linewidth]{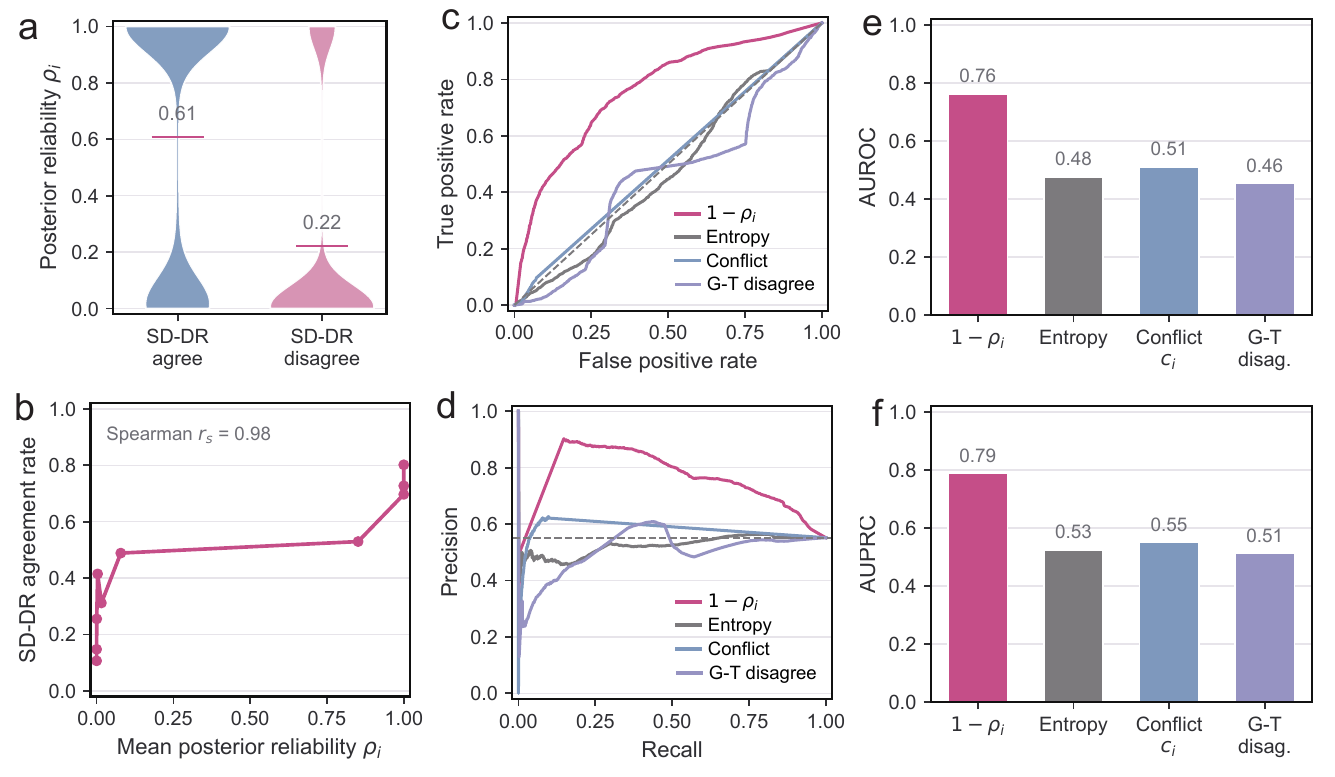}
    \caption{Posterior reliability analysis on paired SD--DR samples. \textbf{a}, Reliability distributions for SD--DR agreement and disagreement. \textbf{b}, SD--DR agreement rate across reliability bins. \textbf{c,d}, ROC and precision--recall curves for detecting SD--DR disagreement. \textbf{e,f}, AUROC and AUPRC summaries.}
    \label{fig:reliability_analysis}
\end{figure}

\begin{figure*}
    \centering
    \includegraphics[width=1.0\linewidth]{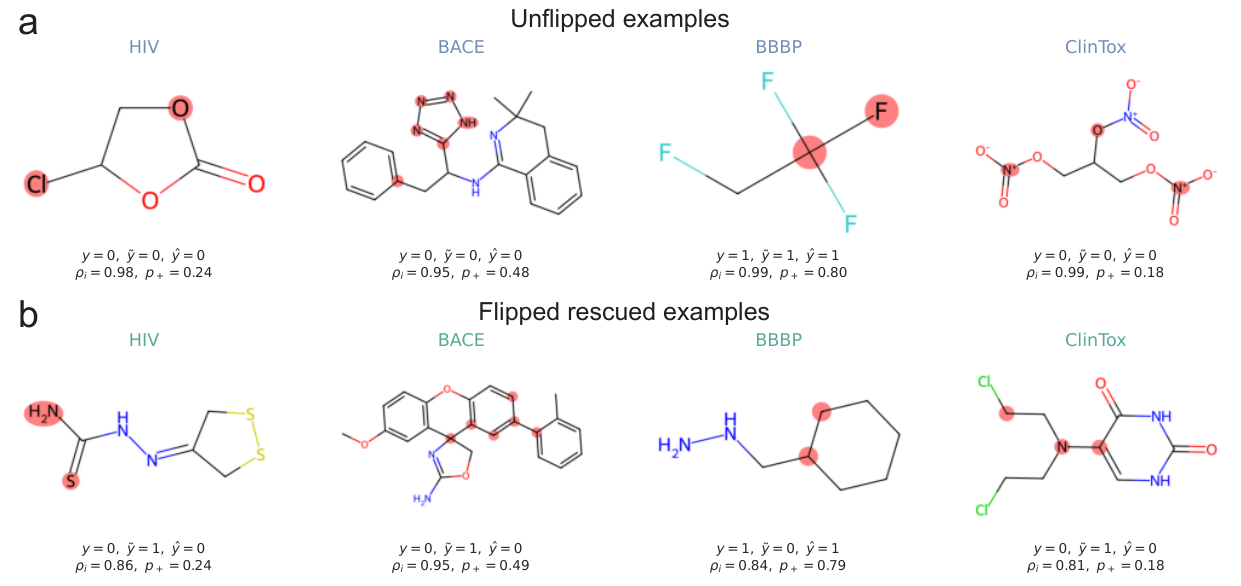}
    \caption{Molecular attribution examples on MoleculeNet benchmarks. \textbf{a}, Unflipped samples where $y$, $\tilde y$, and $\hat y$ agree. \textbf{b}, Flipped samples where $\hat y$ matches $y$ rather than the corrupted $\tilde y$. Highlighted atoms denote model-attributed evidence for the clean prediction.}
    \label{fig:molecular_visualization}
\end{figure*}

To assess whether the posterior reliability derived from the label-observation channel reflects recorded-label quality, we analyze paired SD--DR samples in \datasets{}, where SD labels serve as noisy recorded labels and DR labels provide higher-confidence references. As shown in Figure~\ref{fig:reliability_analysis}a, SD--DR-consistent samples receive substantially higher posterior reliability than inconsistent samples. When samples are grouped by reliability, the empirical SD--DR agreement rate increases monotonically with the mean reliability, yielding a Spearman correlation of $r_s=0.98$ (Figure~\ref{fig:reliability_analysis}b). We further use $1-\rho_i$ as an inconsistency score for detecting SD--DR disagreement and compare it with predictive entropy, evidence conflict $c_i$, and graph--text disagreement. The reliability-based score achieves the strongest ROC and precision--recall performance, with AUROC 0.76 and AUPRC 0.79 (Figure~\ref{fig:reliability_analysis}c--f). These results suggest that $\rho_i$ captures label trustworthiness more effectively than generic uncertainty or cross-modal disagreement alone.

To examine whether the clean posterior can resist corrupted supervision at the molecule level, we visualize representative examples under the 30\% controlled label-flipping setting. For unflipped samples, the clean label $y$, recorded label $\tilde y$, and prediction $\hat y$ are consistent, and the model assigns high posterior reliability (Figure~\ref{fig:molecular_visualization}a). For flipped samples, the recorded label is corrupted, but the selected predictions match the clean label rather than the flipped label (Figure~\ref{fig:molecular_visualization}b). The highlighted atoms indicate model-attributed regions that support the clean prediction. These case studies provide qualitative evidence that \method{} can rely on molecular evidence rather than directly memorizing corrupted labels; however, the highlighted substructures should be interpreted as attribution signals rather than experimentally validated chemical mechanisms.
\vspace{-0.3cm}
\section{Conclusion}

We presented \method{}, a noise-aware framework for learning multimodal molecular representations from noisy labels. \method{} separates latent molecular property prediction from noisy label observation by composing graph and text views as residual evidence for a clean-property posterior and linking it to recorded labels through a categorical label-observation channel. Experiments on naturally noisy and controlled label-flipping benchmarks show strong robustness over representative graph-based, multimodal, and noisy-label learning baselines. The posterior reliability and modality-specific evidence further provide useful diagnostics for interpreting noisy labels and graph--text contributions.

\bibliographystyle{oup-plain}
\bibliography{reference}

\clearpage
\onecolumn
\begin{appendices}

\section{Notation summary}

\begin{center}\scriptsize
\begin{tabularx}{\textwidth}{lY}
\toprule
Symbol & Meaning \\
\midrule
$G_i=(V_i,E_i,X_i)$ & Molecular graph for molecule $i$ with atoms, bonds, and atom features \\
$T_i$ & Text-derived molecular view, descriptor summary, SMILES-derived text, or cached language-model representation \\
$\tilde y_i$, $y_i$ & Recorded noisy class and latent clean molecular property class for molecule $i$ \\
$\mathcal Y$, $C$ & Categorical label space and number of classes \\
$\mathbf z_i^g,\mathbf z_i^t$ & Graph and text molecular representations \\
$\mathbf u_i^g,\mathbf u_i^t$ & Graph- and text-derived residual natural-parameter evidence \\
$\boldsymbol\ell_i$ & Clean categorical logit vector after residual evidence composition \\
$\mathbf p_i$ & Clean-property posterior distribution used for inference \\
$\mathbf O$ & Categorical label-observation matrix linking clean labels to recorded labels \\
$\tilde{\mathbf p}_i$ & Recorded-label distribution after applying the label-observation channel \\
$\rho_i$ & Posterior reliability that the recorded class matches the latent clean class \\
$M_i^g,M_i^t$ & Relative graph and text evidence contributions for diagnostics \\
$c_i$ & Cross-modal evidence-conflict score \\
\bottomrule
\end{tabularx}
\end{center}

\section{Datasets}
\begin{table*}[h]
\centering
\caption{Statistics of the \datasets{} datasets.}
\label{tab:datasets}
\resizebox{0.98\textwidth}{!}{
\begin{tabular}{l l c c c c c c}
\toprule
Task ID & Target & $N$ train/val/test & Pos. rate (SD/DR) & SD-DR disagr. & Avg. atoms & Notes \\
\midrule
NSD2 & NSD2 Methyltransferase & 296,386/694/694 & 0.0\% / 24.9\% & 75.1\% & 25 & High noise, Extreme SD imbalance \\
SKN1 & SKN-1 Transcription Factor & 344,227/663/664 & 0.1\% / 18.4\% & 81.5\% & 24 & High noise, Large SD, Extreme SD imbalance \\
CASP6 & Caspase 6 & 260,564/403/403 & 0.4\% / 49.4\% & 48.3\% & 26 & Extreme SD imbalance \\
RAD52 & RAD52 DNA Repair & 265,560/243/243 & 0.2\% / 73.9\% & 25.5\% & 34 & Low noise, DR imbalanced, Extreme SD imbalance \\
GSK3A & GSK-3$\alpha$ Kinase & 298,513/1023/1023 & 0.1\% / 34.4\% & 29.7\% & 24 & Low noise, Large DR, Extreme SD imbalance \\
GIV & GIV-G$\alpha$ i PPI & 201,917/284/285 & 0.2\% / 37.4\% & 43.4\% & 26 & Extreme SD imbalance \\
UBC13 & UBC13 Ubiquitin Ligase & 312,975/486/487 & 0.1\% / 28.2\% & 71.8\% & 31 & High noise, Large SD, Extreme SD imbalance \\
\bottomrule
\end{tabular}
}
\vspace{-0.3cm}
\end{table*}

\begin{table}[h]
\centering
\caption{Statistics of the MoleculeNet datasets.}
\label{tab:dataset_control}
\resizebox{0.4\textwidth}{!}{
\begin{tabular}{lcccc}
\toprule
\textbf{Dataset} & \textbf{Classes} & \textbf{Graphs} & \textbf{Avg. nodes} & \textbf{Avg. edges} \\
\midrule
HIV     & 2 & 41,127 & 25.5 & 27.5 \\
BACE    & 2 & 1,513  & 34.1 & 36.9 \\
BBBP    & 2 & 2,039  & 23.9 & 25.2 \\
ClinTox & 2 & 1,478  & 26.1 & 27.7 \\
\bottomrule
\end{tabular}
}
\vspace{-0.2cm}
\end{table}

\section{Baselines}
\label{app:baseline}

We compare \method{} with representative baselines from four groups: standard graph neural networks, molecular representation learning methods, multimodal molecular learning methods, and robust graph learning methods under noisy labels. All methods are trained, selected, and evaluated using the same splits and evaluation protocol as \method{}. For MF-PCBA-Noisy7, models are trained on noisy SD labels and evaluated on DR labels; for MoleculeNet, models are trained with flipped training and validation labels and evaluated on clean test labels.

\textbf{GCN.}
Graph Convolutional Network (GCN) is a standard message-passing GNN based on localized graph convolution~\cite{GCN}. It updates atom representations by aggregating normalized neighborhood information and obtains graph-level molecular representations through readout. We include GCN as a basic graph-only baseline.

\textbf{GAT.}
Graph Attention Network (GAT) extends message passing with learnable attention weights over neighboring nodes~\cite{GAT}. This allows the model to emphasize informative atom neighborhoods instead of uniformly aggregating all neighbors. We include GAT as an attention-based graph-only baseline.

\textbf{GIN.}
Graph Isomorphism Network (GIN) is an expressive GNN designed to match the discriminative power of the Weisfeiler--Lehman graph isomorphism test~\cite{gin}. It uses sum aggregation and multilayer perceptrons to capture molecular substructures. We use GIN as a strong graph-only baseline and as the graph encoder backbone for \method{}.

\textbf{GraphCL.}
GraphCL learns graph representations by maximizing agreement between differently augmented views of the same graph~\cite{GraphCL}. It uses graph augmentations such as node dropping, edge perturbation, attribute masking, and subgraph sampling. We include GraphCL as a self-supervised molecular graph representation baseline.

\textbf{GROVE.}
GROVE, also known as GROVER, is a large-scale self-supervised molecular graph Transformer~\cite{GROVE}. It pretrains molecular encoders with node-level, edge-level, and graph-level tasks, including contextual property prediction and motif prediction. We include GROVE as a strong pretrained graph-based molecular representation baseline.

\textbf{SmiSGT.}
SmiSGT, introduced as SimSGT, is a masked graph modeling method for molecular pretraining~\cite{SmiSGT}. It revisits the roles of graph tokenization and decoding, using a Simple GNN-based Tokenizer and remask decoding to improve molecular representation learning. We include it as a substructure-aware self-supervised baseline.

\textbf{Uni-Mol.}
Uni-Mol is a universal 3D molecular representation learning framework pretrained on large-scale molecular conformations and protein pocket structures~\cite{uni-mol}. It incorporates 3D atomic coordinates through a Transformer-based architecture and supports both property prediction and 3D molecular tasks. We include Uni-Mol as a pretrained 3D molecular baseline.

\textbf{S-GCIB.}
S-GCIB is a subgraph-conditioned graph information bottleneck method for molecular graph pretraining~\cite{s-gcib}. It learns graph cores and significant subgraphs through graph compression, subgraph extraction, and attention-based interaction. We include S-GCIB as a subgraph-aware molecular pretraining baseline.

\textbf{Tri-SGD.}
Tri-SGD is a triple-modal molecular fusion baseline derived from multimodal fused deep learning for drug property prediction~\cite{Tri-SGD}. It processes SMILES-encoded vectors, ECFP fingerprints, and molecular graphs with Transformer, BiGRU, and GCN modules, respectively. We include Tri-SGD as a representative late-fusion multimodal baseline.

\textbf{MMSG.}
MMSG jointly learns molecular representations from SMILES strings and molecular graphs~\cite{MMSG}. It introduces graph-derived bond-level information as an attention bias in the SMILES Transformer and uses a bidirectional message communication GNN to enhance graph representations. We include MMSG as a graph--SMILES multimodal baseline.

\textbf{MDFCL.}
MDFCL is a multimodal graph contrastive learning framework for molecular property prediction~\cite{mdfcl}. It integrates graph and sequence modalities and designs adaptive molecular augmentations based on backbones and side chains. We include MDFCL as a multimodal contrastive learning baseline.

\textbf{ProtoMol.}
ProtoMol is a prototype-guided multimodal molecular learning method~\cite{protomol}. It models molecular graphs and textual descriptions with dual-branch encoders, cross-modal attention, and a shared prototype space. We include ProtoMol as a recent graph--text multimodal baseline.

\textbf{OMG.}
OMG is a robust graph classification method for learning with noisy graph labels~\cite{omg}. It combines graph contrastive learning with coupled Mixup and uses neighbor-aware noise removal to reduce the impact of unreliable labels. We include OMG as a representative noisy-label graph learning baseline.

\textbf{SPORT.}
SPORT addresses noisy graph classification from a subgraph perspective~\cite{sport}. It represents each graph as a set of perturbed subgraphs, encodes them with an equivariant network, and updates potentially noisy labels using subgraph-level predictions. We include SPORT as a subgraph-based robust learning baseline.

\textbf{RTGNN.}
RTGNN is a robust training framework for GNNs under scarce and noisy labels~\cite{rtgnn}. It separates clean and noisy label candidates, applies self-reinforcement for label correction, and uses consistency regularization to prevent overfitting. We include RTGNN as a noise-governance graph learning baseline.

\textbf{TFR.}
TFR, or Topological Feature Reconstruction, mitigates label noise by using graph topology and feature reconstruction~\cite{TFR}. It assumes that clean label patterns share more information with graph structure and node features than corrupted labels. We include TFR as a topology-guided noisy-label graph learning baseline.



\end{appendices}

\end{document}